\documentclass[a4paper,11pt,twocolumn,twoside]{article}
\usepackage{graphicx}
\usepackage{sepln_en}
\usepackage{fullname}
\usepackage[utf8]{inputenc}
\usepackage[english]{babel}
\usepackage{multirow}
\usepackage{booktabs}

\input epsf

\title{AI-generated Text Detection with a GLTR-based Approach}

\author {\textbf{Lucía Yan Wu,} \textbf{Isabel Segura-Bedmar}\\
Universidad Carlos III de Madrid, Leganés, Madrid\\
lyluciayan@gmail.com, isegura@inf.uc3m.es\\
}

\seplntranstitle{Detección de Textos Generados por IA con un enfoque basado en GLTR}

\seplnkey{Machine-Generated Text, GPT-2, GLTR, Text Classification.}

\seplnabstract{The rise of LLMs (Large Language Models) has contributed to the improved performance and development of cutting-edge NLP applications.
However, these can also pose risks when used maliciously, such as spreading fake news, harmful content, impersonating individuals, or facilitating school plagiarism, among others. This is because LLMs can generate high-quality texts, which are challenging to differentiate from those written by humans. 
GLTR, which stands for Giant Language Model Test Room and was developed jointly by the MIT-IBM Watson AI Lab and HarvardNLP, is a visual tool designed to help detect machine-generated texts based on GPT-2, that highlights the words in text depending on the probability that they were machine-generated. One limitation of GLTR is that the results it returns can sometimes be ambiguous and lead to confusion. This study aims to explore various ways to improve GLTR's effectiveness for detecting AI-generated texts within the context of the IberLef-AuTexTification 2023 shared task, in both English and Spanish languages.
Experiment results show that our GLTR-based GPT-2 model overcomes the state-of-the-art models on the English dataset with a macro F1-score of 80.19\%, except for the first ranking model (80.91\%). However, for the Spanish dataset, we obtained a macro F1-score of 66.20\%, which differs by 4.57\% compared to the top-performing model.
}

\seplnclave{Texto generado por IA, GPT-2, GLTR, Clasificación de texto.}

\seplnresumen{El auge de los LLMs (Modelos de Lenguaje de Gran Tamaño) ha contribuido al mejor desempeño y desarrollo de aplicaciones de PLN de vanguardia. Sin embargo, estos modelos también pueden suponer riesgos cuando se utilizan de manera maliciosa, como la difusión de noticias falsas o contenido dañino, la suplantación de identidad o la facilitación del plagio escolar, entre otros. Esto se debe a que los LLMs pueden generar textos de alta calidad, los cuales son difíciles de diferenciar de aquellos escritos por humanos. 
GLTR, que significa Giant Language Model Test Room y fue desarrollado conjuntamente por el MIT-IBM Watson AI Lab y HarvardNLP, es una herramienta visual diseñada para ayudar a detectar textos generados por máquinas basados en GPT-2, que resalta las palabras en el texto dependiendo de la probabilidad de que hayan sido generadas de forma automática. Una limitación de GLTR es que los resultados que devuelve pueden, en ocasiones, ser ambiguos y llevar a confusión. Este estudio tiene como objetivo explorar diversas formas de mejorar la eficacia de GLTR para detectar textos generados por IA en el contexto de la tarea compartida IberLef-AuTexTification 2023, tanto en inglés como en español.
Los resultados muestran que nuestro modelo GPT-2 basado en GLTR obtiene una puntuación macro F1 de 80.19\% para la tarea en inglés. Nuestro enfoque supera a todos los sistemas que participaron en la tarea, excepto al primer sistema, que obtuvo una puntuación macro F1 ligeramente superior a la nuestra (80.91\%). Sin embargo, para el conjunto de datos en español, obtuvimos una puntuación macro F1 de 66.20\%, lo que representa una diferencia del 4.57\% respecto al modelo con mejor desempeño.}

\firstpageno{1}

\begin{document}


\setlength\titlebox{25cm} 

\label{firstpage} \maketitle

%

\section{Introduction}

In recent years, the rapid development of AI (Artificial Intelligence) due to the release of models such as ChatGPT by OpenAI or Gemini by Google, has increased the amount of AI-generated content on a huge scale \cite{martínez2023combining}. GPT \cite{openai2024gpt4}, PaLM \cite{chowdhery2022palm}, and BLOOM \cite{workshop2023bloom} are some of the Large Language Models (LLMs) accessible to the public, applied for automatic content generation. 

LLMs are known for being able to generate high-quality texts, i.e., texts that are grammatically correct and coherent, which can make it challenging to differentiate them from those written by humans \cite{llm-risks-bublak}. Nevertheless, even though LLMs have achieved remarkable success in the field, they have also introduced some risks and ethical concerns. This is because LLMs do not understand the meaning of the content they generate, and hence can produce false or inaccurate information, leading to hallucinations \cite{huang2023survey}. In addition, bias can be introduced during the generation of content, leading to unethical content such as gender inequality \cite{unesco-llm-bias}. Furthermore, the misuse of these models by some users may contribute to the spread of fake news, spam for malicious purposes, or even enhancing academic cheating \cite{llm-risks-upwork} \cite{llm-risks-heldt}.

For these reasons, it is crucial to develop some approaches to regulate and detect AI-generated content. To do so, the AuTexTification \cite{autextification} shared task was presented as part of the IberLEF 2023 workshop \cite{jimenez2023overview}. The task involved two subtasks: binary classification of text as human-written or AI-generated, and identification of the specific LLM that generated the text.

In this study, we introduce a new approach using the GLTR tool \cite{gehrmann-etal-2019-gltr} for tackling the first subtask in both English and Spanish. GLTR (Giant Language Model Test Room) is a visual tool designed to help detect machine-generated texts based on GPT-2, that highlights the words in text depending on the probability that they were machine-generated. However, one limitation of GLTR is that the results it returns can sometimes be ambiguous and lead to confusion. Hence, this study aims to explore various ways to improve GLTR's effectiveness for the binary classification task.

The rest of this paper is organized as follows: 
In Section 2 we report the most relevant state-of-the-art approaches for AI-generated text detection. In Section 3, we present the datasets used in this study, and in Section 4, we describe the experimental setup considered to asses the binary classification task. In Section 5, we show and discuss the results obtained from the proposed method in both languages, English and Spanish. Finally, Section 6 presents the conclusions and future work.

\section{State of the art}
\subsection{Participating systems in AuTexTification 2023} \label{sec:autextification2023}

\subsubsection{English}
In the IberLef-AuTexTification shared task 2023 \cite{autextification}, 36 teams participated in Subtask 1 for English texts. Most submissions utilized transformer-based models, especially BERT-based models, while others employed generative models (such as GPT-2, Grover, and OPT), CNNs, or LSTMs. Traditional machine learning models, such as Logistic Regression and SVM, were also implemented by participants but generally performed worse compared to transformer-based models.

The top-performing team, TALN-UPF \cite{taln-upf2023bilstm}, achieved a macro F1-score of 80.91\% using a bidirectional LSTM with fine-tuned \texttt{RoBERTa} and token-level POS tagging, which helped detect grammatical and morphological errors uncommon for humans. This same team also placed second with a macro F1-score of 74.16\% using the same model without POS tagging. The third-ranked team, CIC-IPN-CsCog \cite{cic-ipn-cscog2023gpt}, used a fine-tuned GPT-2, achieving a macro F1-score of 74.13\%.

Traditional machine learning models achieved lower scores; for instance, Lingüística\_UCM's \cite{linguistica-uc3m2023linearsvc} LinearSVC with TF-IDF and n-gram features reached 68.33\%, while other models like CatBoost and Multilayer Perceptron performed below 60\%. 

Overall, transformer and neural network models outperformed traditional machine learning approaches, with BiLSTM models leading in effectiveness, followed by GPT-2 and other ensemble transformer-based models.

\subsubsection{Spanish}
In Subtask 1 for Spanish texts, a total of 23 teams participated. Similarly to the English task, most submissions were based on transformer models, obtaining better results compared to other teams that used more traditional approaches. Moreover, the macro F1-scores obtained were lower for all of the models compared to their performance in the English task. This could be attributed to the limited number of models trained on data in the Spanish language.

The best-performing team, as well as in the English task, was TALN-UPF \cite{taln-upf2023bilstm}, with their BiLSTM model with fine-tuned \texttt{RoBERTa} and token-level POS tagging, obtaining a macro F1-score of 70.77\%. Lingüística\_UCM \cite{linguistica-uc3m2023linearsvc} achieved second place with their LinearSVC with TF-IDF and n-gram features with a score of 70.60\%.

Overall, we can support that indeed solutions based on transformer models and neural network algorithms give the best performance, overbeating all of the proposed baselines in the task.
On the other hand, traditional Machine Learning models such as SVM or CatBoost classifier tend to return lower scores.

\subsection{IberAuTexTification 2024}
In 2024, IberLef-IberAuTexTification shared task \cite{PLN6628}, the second version of the AuTexTification at IberLEF 2023 shared task, was released. It was based on the same subtasks as the 2023 edition, but focused on languages from the Iberian Peninsula: Spanish, Catalan, Basque, Galician, Portuguese, and English (in Gibraltar). Additionally, new domains (news, reviews, emails, essays, dialogues, wikipedia, wikihow, tweets, etc.), and models (GPT, LLaMA, Mistral, Cohere, Anthropic, MPT, Falcon, etc.) were introduced for the generation of automatically generated texts.

The top-performing systems for both subtasks were Transformer models enhanced with additional lexical, syntactic, and semantic features.
The best-performing team, jor\_isa\_uc3m \cite{jor_isa_uc3m}, proposed an ensemble composed by three multilingual transformers, \texttt{DistilBERT-base-multilingual-cased}, \texttt{mDeBERTa-v3-base}, and \texttt{XLM-RoBERTa-base}, obtaining a macro F1-score of 80.5\%.

\subsection{Other approaches to detect AI-generated texts}
In April 2024, Fareed Khan published a web application\footnote{https://ai-text-detect-easy.streamlit.app/} that identifies the 100 most common words used by AI within an input text. The objective of this application was to detect if the text was generated by a machine just by looking at it.

GPTZero\footnote{https://gptzero.me/}\cite{tian2023gptzero} launched their first AI detection solution in January 2023, and has since then developed several tools for this goal. Among them is an API available online that allows the user to input a text, and returns the probability percentage for each class: human, mixed, and AI, along with the confidence percentage. 

In 2019, researchers at Harvard and IBM \cite{gehrmann-etal-2019-gltr} developed a visual tool named GLTR (Giant Language Model Test Room), which highlighted the words in a text depending on how predictable or common these were, i.e., the probability these were generated by a machine. 
The words were colored by order of frequency, in purple, red, yellow, or green, where purple represented the rarest words with a predicted position higher than 1,000, and green the most common (within the top 10 predictable words). This tool was tested in an assessment where human subjects initially identified AI-generated texts manually with an accuracy of 54\%, which increased to over 72\% after using GLTR.

Generally, we can see that there are various ways of detecting AI-generated texts, going from manual extraction to using LLMs and other visual tools. We tested each of the approaches defined above, where Fareed Khan's app seemed inconsistent since it produced similar lists of words for both human and AI-generated texts. Next, GPTZero always returned the correct label, differentiating very well the texts. Lastly, the GLTR tool gave in general good results but required user interpretation, which sometimes led to confusion.

\section{Dataset}
The dataset used for the study belongs to the AuTexTification shared task of the IberLEF 2023 Workshop \cite{autextification}. This dataset is divided by subtask and language, and by train or test. This results in eight different datasets, which collectively contain more than 160,000 texts across five domains with diverse writing styles ranging from more structured and formal to less structured and informal. For subtask 1, the training split included tweets, how-to articles, and legal documents, while the test split consisted of reviews and news articles. This setup was designed to assess the model's ability to generalize to different text domains.


The source datasets used for extracting the human-authored texts can be seen in Table \ref{tab:source-human-texts} below:
\begin{table}[!h]
    \centering
    \resizebox{\columnwidth}{!}{
    \begin{tabular}{c|cc}
      {\bf } &{\bf English} &{\bf Spanish}\\
      \hline
      \textbf{Legal} &\textit{MultiEURLEX} &\textit{MultiEURLEX}\\
      \textbf{News} &\textit{XSUM} &\textit{MLSUM \& XLSUM}\\ 
      \textbf{Reviews} &\textit{Amazon Reviews} &\textit{COAR \& COAH}\\   
      \textbf{Tweets} &\textit{TSATC}
      &\textit{XLM-Tweets \& TSD}\\
      \textbf{How-to} &\textit{WikiLingua} &\textit{WikiLingua}\\
    \end{tabular}
    }
    \caption{Human-authored source datasets.}
    \label{tab:source-human-texts}
\end{table}

Machine-generated texts were created from human texts by using three different BLOOM models (\texttt{BLOOM-1B7}, \texttt{BLOOM-3B}, and \texttt{BLOOM-7B1}), and three GPT-3 models (\texttt{babbage}, \texttt{curie}, and \texttt{text-davinci-003}, with 1b, 6.7b and 175b parameter scales respectively). These models were fine-tuned with a top-p of 0.9 and a temperature of 0.7.

Moreover, a maximum number of tokens was selected for each domain, maintaining the distribution of human texts: 20 tokens for tweets, 70 for reviews, and 100 for news, legal, and how-to articles. 

\begin{figure}[!h]
 \centering
 \includegraphics[width=7cm,clip]{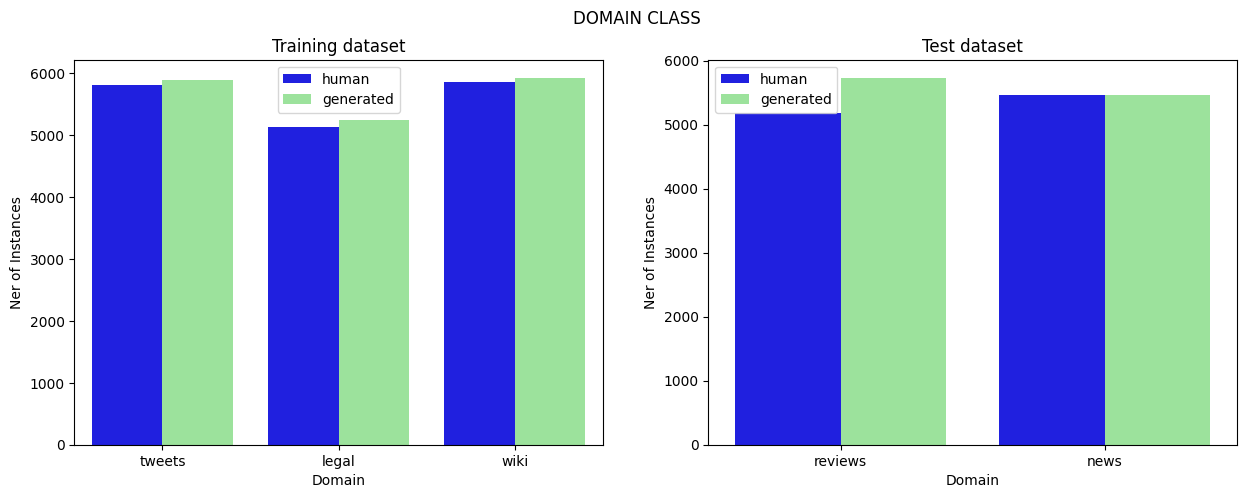}  
 \caption{\centering Number of samples per dataset, class, and domain (subtask 1 for English).}
 \label{fig:balanced_en}
\end{figure}

\begin{figure}[!h]
 \centering
 \includegraphics[width=7cm,clip]{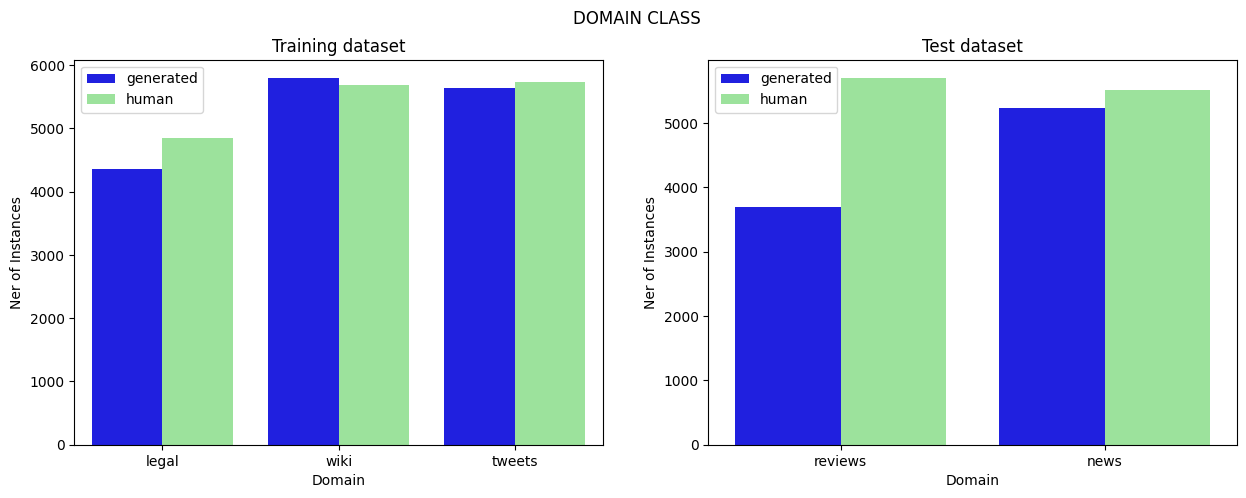}
 \caption{\centering Number of samples per dataset, class, and domain (subtask 1 for Spanish).}
 \label{fig:balanced_es}
\end{figure}

Table \ref{tab:distribution}, displays the resulting dataset distribution. We observe that it is well-balanced for both domains and classes, for both languages. Moreover, Figures \ref{fig:balanced_en} and \ref{fig:balanced_es} support this statement. However, we can see that the reviews domain in the Spanish language, contains slightly fewer generated instances than human instances.
\begin{table}
  \centering
  \begin{tabular}{lllccc}
    \toprule
    & Domain & GEN & HUM & $\Sigma$ \\
    \midrule
    \multirow{5}{*}{\rotatebox[origin=c]{90}{English}} 
    & Tweets & 5,813 & 5,884 & 11,697 \\
    & How-to & 5,862 & 5,918 & 11,780 \\
    & Legal & 5,124 & 5,244 & 10,368 \\
    \cmidrule{2-5}
    & News & 5,464 & 5,464 & 10,928 \\
    & Reviews & 5,726 & 5,178 & 10,904 \\
    \midrule
    & Total & 27,989 & 27,688 & 55,677 \\
    \midrule
    \midrule
    \multirow{5}{*}{\rotatebox[origin=c]{90}{Spanish}} 
    & Tweets & 5,739 & 5,634 & 11,373 \\
    & How-to & 5,690 & 5,795 & 11,485 \\
    & Legal & 4,846 & 4,358 & 9,204 \\
    \cmidrule{2-5}
    & News & 5,514 & 5,223 & 10,737 \\
    & Reviews & 5,695 & 3,697 & 9,392 \\
    \midrule
    & Total & 27,484 & 24,707 & 52,191 \\
    \bottomrule
  \end{tabular}
  \caption{Statistics of the datasets for subtask 1.}
  \label{tab:distribution}
\end{table}


\section{Methodology}

In this work, we present a novel approach that combines the GLTR tool, which visualizes the likelihood of word predictions, with the predictive capabilities of GPT-2 models to classify text as either human-written or machine-generated. By leveraging GLTR's word-level probabilities and extending its functionality with a classification step, we created a robust pipeline for evaluating text authenticity. This pipeline is applied to English and Spanish texts, using multiple versions of GPT-2.

\begin{figure} 
    \centering
    \includegraphics[width=7cm]{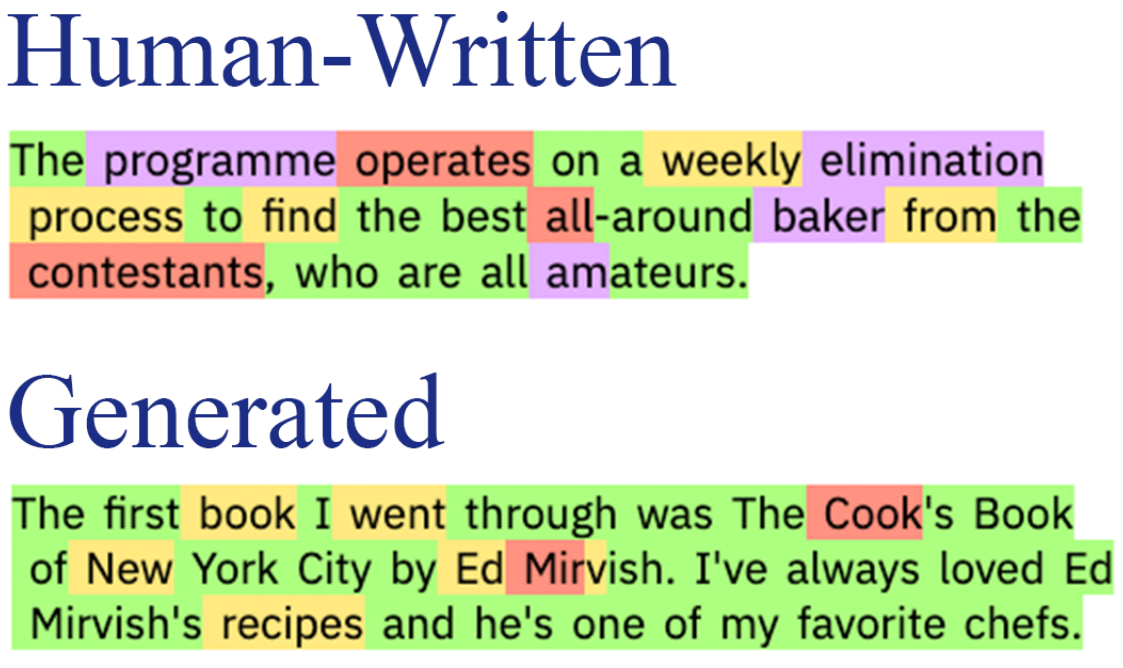}
    \caption{\centering GLTR example \cite{DBLP:journals/corr/abs-1906-04043}.}
    \label{fig:gltr-example}
\end{figure}

GLTR, also called Giant Language Model Test Room, is a visual tool developed by researchers at Harvard NLP and MIT-IBM Watson AI lab, which uses GPT-2 to predict the likelihood of words in a sentence \cite{DBLP:journals/corr/abs-1906-04043}. The model iterates through the different words of an input sequence, where for each word it returns a set of predicted words along with their probability of being the next word. If the real word is contained within the Top 10 predicted words, it is colored green, and the same happens for the Top 100 in yellow, the Top 1000 in red, and the rest in purple. We can better understand this by looking at the example in Figure \ref{fig:gltr-example}, where on top there is a human-written text, and below a machine-generated text. We can see that the generated text contains for the most part green-colored words, while the human-written text is composed of a variety of colored words. This is because humans tend to use more complex words and machines, more common and predictable words. Moreover, the contributors of this tool also made available a demo\footnote{http://demo.gltr.io/client/index.html} for users to test.

As we can see, GLTR is a very remarkable tool, and hence, we decided to create a model based on it. We used the same working structure of the GLTR, taking as reference the code in the GitHub repository\footnote{https://github.com/HendrikStrobelt/detecting-fake-text} that the contributors have made available. 

To construct the GLTR-based model, we first preprocessed the dataset by splitting it into training (80\% of instances) and validation (20\% of instances) subsets, and encoding the labels \textit{generated} and \textit{human} as 0 and 1, respectively. Then, each sentence was tokenized using GPT-2's tokenizer, and predictions for word likelihoods were obtained for every word. This was done by defining a function to obtain the ordered list of predicted words and their probabilities, given the previous words in the sentence. Given this list, we found the position of the ground-truth word and colored it according to the rules we explained before.

What differentiates our model from the GLTR tool is that we have incorporated an additional evaluation step to compute some performance metrics. While GLTR primarily focuses on visualizing the likelihood of words being generated by a model, we take a more data-driven approach by classifying sentences based on the proportion of green-colored words, which are those predicted to have a higher likelihood of being machine-generated.
In our approach, the proportion of green-colored words is calculated for each sentence and compared against a predefined threshold to classify the sentence. Specifically, if the percentage of green-colored words exceeds the threshold, the sentence is classified as machine-generated; if it is lower, it is classified as human-written.
For example, if the threshold is set at two-thirds, the sentence is classified as machine-generated when more than two-thirds of its words are green-colored. Otherwise, it is labelled as human-written.

In our study, we experimented with various threshold values (1/4, 1/3, 1/2, 2/3, 3/4, and 5/6) to explore the impact on classification performance across both English and Spanish languages. These variations allowed us to assess how the proportion of machine-like words influences classification accuracy and fine-tune the model's ability to distinguish between human-written and machine-generated texts.

Once the classification is made, the predicted classes are then compared to the real labels, and the macro F1-scores are computed.

The GPT-2 model is a pretrained model on large amounts of raw and unlabelled texts in the English language, from the internet. The model was developed by OpenAI, and specifically designed to guess the next word in a sentence. During training, inputs consist of continuous text sequences, where the model learns to predict the next word in the sequence by shifting the target sequence one token to the right.
Moreover, the model uses a mask-mechanism to ensure that each prediction is based solely on previous tokens, preventing the model from using future ones. This approach enables the model to excel at generating text from prompts.
However, one limitation of GPT-2 is that the training data used for this model contains a lot of unfiltered content from the internet, which can introduce biases into the model.

For the English texts, we studied 4 different versions of the GPT-2 model \cite{radford2019language}: 
\begin{itemize}
    \item \texttt{gpt2-small},is the smallest version of GPT-2, with 124M parameters.
    
    \item \texttt{gpt2-medium}, is the 355M parameter version.
    \item \texttt{gpt2-large}, is the 774M parameter version.
    \item \texttt{gpt2-xl}, is the 1.5B parameter version.
\end{itemize}

For the Spanish texts, we additionally used the \texttt{gpt2-small-spanish} model developed by Datificate \cite{datificate}, which is based on the \texttt{gpt2-small} model \cite{radford2019language}.
It was trained on Spanish Wikipedia (around 3GB of processed training data), using Transfer Learning and Fine-tuning techniques. Similarly to the English pre-trained \texttt{gpt2-small}, one limitation of this model is the introduction of biases into the model, originated from the unfiltered content from the internet used as training data.

Note that the study with different threshold values was done on \texttt{gpt2-small} and \texttt{gpt2-small-spanish} for English and Spanish, respectively. To prevent computational complexity, \texttt{gpt2-medium}, \texttt{gpt2-large}, and \texttt{gpt2-xl}, were only evaluated on the best threshold value obtained on the \texttt{gpt2-small} and \texttt{gpt2-small-spanish} study.

The dataset as well as the code to replicate the experiments can be found in the GitHub repository \footnote{https://github.com/xxxxxxx/AI-generated-Text-Detection-with-GLTR-based-approach}. Moreover, an online demo has been deployed in Streamlit, which is available to all users and allows them to test their own English texts, returning if the input text was AI-generated or human-written \footnote{https://ai-generated-text-detection-with-gltr-based-approach.streamlit.app/}.

\section{Results and discussion}
For our study, we mainly focused on the macro F1-score and the F1-scores for each of the classes, to compare the results between the different models. The reason is that the AuTexTification task on which this project is based used the F1-score as the main performance metric.

Table \ref{tab:gpt2-thresholds-eval} shows the results obtained from the experimentation of the \texttt{gpt2-small} model under various threshold values. 
The best score for the Generated class is achieved at the threshold of 2/3, with an F1-score of 82.49\%. While for the Human class, it is obtained at the threshold of 3/4, with an F1-score of 79.30\%.
Overall, the most optimal threshold value is 2/3, which returns the highest macro F1-score, 80,19\%, by balancing the performance between the Generated and Human categories. 
\begin{table}[!h]
    \centering
    \resizebox{\columnwidth}{!}{
    \begin{tabular}{ccccc}  
      \hline
      {\bf Threshold} &{\bf Generated} &{\bf Human} &{\bf Macro avg F1}\\
      \hline
      \textbf{\texttt{1/4}} &0.6786 &0.0075 &0.3430\\
      \hline
      \textbf{\texttt{1/3}} &0.6798 &0.0199 &0.3498\\ 
      \hline
      \textbf{\texttt{1/2}} &0.6982 &0.1784 &0.4383\\
      \hline
      \textbf{\texttt{2/3}} &0.8249 &0.7790 &0.8019\\
      \hline
      \textbf{\texttt{3/4}} &0.7029 &0.7930 &0.7480\\
      \hline
      \textbf{\texttt{5/6}} &0.2238 &0.6839 &0.4538\\
      \hline
    \end{tabular}
    }
    \caption{GPT-2 Small F1-scores with different threshold values (subtask 1 for English).}
    \label{tab:gpt2-thresholds-eval}
\end{table}

Next, Table \ref{tab:gpt2-models-eval} represents the scores obtained from the different GPT-2 models with a threshold value of 2/3, experimented for English texts in subtask 1. We can notice that as the model size increases, the performance generally declines across all metrics: Generated F1-score, Human F1-score, and macro F1-score.  Hence, the best results are obtained with the \texttt{gpt2-small} model, with Generated F1-score, Human F1-score, and macro F1-score of 82.49\%, 77.90\%, and 80.19\%, respectively.

\begin{table}[!h]
    \centering
    \resizebox{\columnwidth}{!}{
    \begin{tabular}{lcccc}
      \hline
      {\bf } &{\bf Generated} &{\bf Human} &{\bf Macro avg F1}\\
      \hline
      \textbf{\texttt{gpt2-small}} &0.8249 &0.7790 &0.8019\\
      \hline
      \textbf{\texttt{gpt2-medium}} &0.7963 &0.6777 &0.7370\\ 
      \hline
      \textbf{\texttt{gpt2-large}} &0.7771 &0.6140 &0.6956\\
      \hline
      \textbf{\texttt{gpt2-xl}} &0.7619 &0.5625 &0.6622\\
      \hline
    \end{tabular}
    }
    \caption{GPT-2 models with threshold 2/3 F1-scores (subtask 1 for English).}
    \label{tab:gpt2-models-eval}
\end{table}

\begin{figure}
    \centering
    \includegraphics[width=7cm,clip]{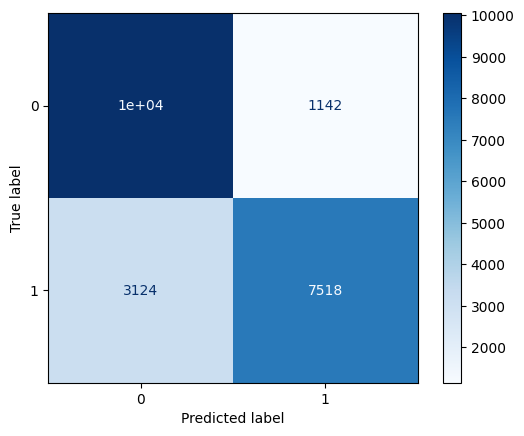}
    \caption{GPT-2 Small Confusion Matrix (0:
generated, 1: human).}
    \label{fig:confusion_matrix_gpt2small_en}
\end{figure}

Figure \ref{fig:confusion_matrix_gpt2small_en}, describes the confusion matrix for \texttt{gpt2-small}, which gives an overview of how well the model classifies generated and human-written texts. Out of 11,190 generated texts, the model correctly classifies 10,048 as generated (true positives). However, it misidentifies 1,142 as human-written (false positives). On the other hand, for the 10,642 human-written texts, the model classifies 7,518 as human, while misclassifying 3,124 as generated (false negatives). Although overall the model performs well, it still has room for improvement in distinguishing between generated and human-written texts, especially in reducing false positives and false negatives.

For the Spanish texts, Table \ref{tab:gpt2-thresholds-eval-es}, presents the F1-scores obtained for different threshold values by the \texttt{gpt2-small-spanish} model. The best Generated F1-score is obtained at the 1/2 threshold with 74.28\%, while the highest Human F1-score at the threshold 2/3 with 68.31\%.
Similarly to \texttt{gpt2-small} for the English dataset, the optimal threshold that achieves the overall balance and highest macro F1-score of 62.18\%, is 2/3.
\begin{table}[!h]
    \centering
    \resizebox{\columnwidth}{!}{
    \begin{tabular}{ccccc}
      \hline
      {\bf Threshold} &{\bf Generated} &{\bf Human} &{\bf Macro avg F1}\\
      \hline
      \textbf{\texttt{1/4}} &0.6788 &0.0242 &0.3515\\
      \hline
      \textbf{\texttt{1/3}} &0.7208 &0.0768 &0.3988\\ 
      \hline
      \textbf{\texttt{1/2}} &0.7428 &0.4458 &0.5943\\
      \hline
      \textbf{\texttt{2/3}} &0.5605 &0.6831 &0.6218\\
      \hline
      \textbf{\texttt{3/4}} &0.2130 &0.6414 &0.4272\\
      \hline
      \textbf{\texttt{5/6}} &0.0320 &0.6180 &0.3250\\
      \hline
    \end{tabular}
    }
    \caption{GPT-2 Small Spanish F1-scores with different threshold values (subtask 1 for Spanish).}
    \label{tab:gpt2-thresholds-eval-es}
\end{table}

Next, Table \ref{tab:es-gpt2-models-eval} shows the results obtained for each of the different GPT-2 models with the threshold value at 2/3. The best macro F1-score is achieved with \texttt{gpt2-xl} with 66.20\%, and with Generated and Human F1-scores of 64.90\% and 67.51\%, respectively. Moreover, we notice that \texttt{gpt2-small-spanish} performs worse regardless of it being trained with Spanish Wikipedia. This could be due to \texttt{gpt2-small-spanish} being much smaller in terms of the number of parameters.
\begin{table}[!h]
    \centering
    \resizebox{\columnwidth}{!}{
    \begin{tabular}{lcccc}
      \hline
      {\bf } &{\bf Generated} &{\bf Human} &{\bf Macro avg F1}\\
      \hline
      \textbf{\texttt{gpt2-small}} &0.0916 &0.6257 &0.3586\\
      \hline
      \textbf{\texttt{gpt2-small-spanish}} &0.5605 &0.6831 &0.6218\\
      \hline
      \textbf{\texttt{gpt2-medium}} &0.3486 &0.6599 &0.5042\\ 
      \hline
      \textbf{\texttt{gpt2-large}} &0.5265 &0.6772 &0.6019\\
      \hline
      \textbf{\texttt{gpt2-xl}} &0.6490 &0.6751 &0.6620\\
      \hline
    \end{tabular}
    }
    \caption{GPT-2 models with threshold 2/3 F1-scores (subtask 1 for Spanish).}
    \label{tab:es-gpt2-models-eval}
\end{table}

\begin{figure}
    \centering
    \includegraphics[width=7cm,clip]{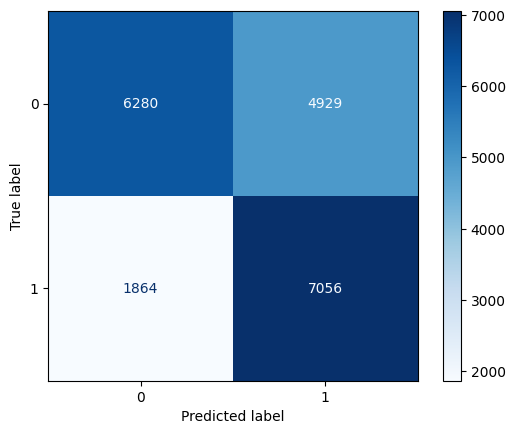}
    \caption{GPT-2 XL Confusion Matrix (0:
generated, 1: human).}
    \label{fig:confusion_matrix_gpt2xl_spanish}
\end{figure}

Figure \ref{fig:confusion_matrix_gpt2xl_spanish}, presents the confusion matrix for \texttt{gpt2-xl} for Spanish texts. Out of 11,209 generated texts, the model correctly identifies 6,280 as generated, but it mistakenly classifies 4,929 as human-written (false positives). For the 8,920 human-written texts, it accurately labels 7,056 as human, while 1,864 are incorrectly classified as generated (false negatives). Similarly to \texttt{gpt2-small} for English texts, there is some room for improvement in reducing the false positives and false negatives.

Moreover, we compare our models' results with those presented in AuTexTification 2023 \cite{autextification}, described in Section \ref{sec:autextification2023}. \texttt{gpt2-small} overbeats all of the presented models except for the BiLSTM with fine-tuned RoBERTa and token-level POS tagging from team TALN-UPF \cite{taln-upf2023bilstm}, achieving second rank. However, we must remark the performance between models is very similar, with only a macro F1-score difference of 0.72\%.
On the other hand, \texttt{gpt2-xl} did not perform so well for Spanish texts, as it would have only ranked 9th position in the competition, due to the low F1-scores obtained for both classes.


\section{Conclusions and future work}
In this study, we evaluated the performance of GPT-2 models extended with the GLTR tool, on the binary classification task of differentiating AI-generated and human-written texts.

For both English and Spanish texts, the optimal threshold for \texttt{gpt2-small} and \texttt{gpt2-small-spanish} was 2/3, achieving a macro F1-score of 80.19\% and 62.18\%, respectively. However, although \texttt{gpt2-xl} was not trained with Spanish data, it returned better results, achieving a macro F1-score of 66.20\%. 

Among the four GPT-2 model variants explored, \texttt{gpt2-small} outperformed larger models (\texttt{gpt2-medium, gpt2-large, gpt2-xl}) in the English subtask. On the other hand, for the Spanish texts, the \texttt{gpt2-xl} model outperformed smaller models (\texttt{gpt2-small, gpt2-small-spanish, gpt2-medium, gpt2-large}).

Moreover, the \texttt{gpt2-small} model would have ranked second in the AuTexTification 2023 \cite{autextification} competition, outperforming most models with a minimal macro F1-score difference of 0.72\% compared to the top-performing BiLSTM + \texttt{RoBERTa} + POS tagging model \cite{taln-upf2023bilstm}. However, for Spanish texts, the \texttt{gpt2-xl} model would have ranked lower, in 9th position, as it returned a low F1-score for both Generated and Human categories.

As for future work, we plan on doing further fine-tuning to improve the \texttt{gpt2-small-spanish} model performance, as well as training the \texttt{gpt2-xl} model with Spanish data. We also plan on tackling the multilingual task for languages from the Iberian Peninsula presented on the IberAuTexTification 2024 task \cite{PLN6628}, as well as the second subtask about model attribution, which is a multiclass classification problem, proposed in the two editions of AuTexTification.
Furthermore, we also plan to investigate the detection of AI-generated content in a multimodal setting.


\bibliographystyle{fullname}
\bibliography{EjemploARTsepln}

\begin{thebibliography}{}

\bibitem[\protect\citename{Aguilar-Canto \bgroup et al.\egroup }2023]{cic-ipn-cscog2023gpt}
Aguilar-Canto, F., M.~Cardoso-Moreno, D.~Jim{\'e}nez, and H.~Calvo.
\newblock 2023.
\newblock Gpt-2 versus gpt-3 and bloom: Llms for llms generative text detection.
\newblock In {\em Proceedings of IberLEF 2023}. CEUR Workshop Proceedings.

\bibitem[\protect\citename{Alonso~Sim{\'o}n \bgroup et al.\egroup }2023]{linguistica-uc3m2023linearsvc}
Alonso~Sim{\'o}n, L., J.~A. Gonzalo~Gimeno, A.~M. Fern{\'a}ndez-Pampill{\'o}n~Cesteros, M.~Fernández~Trinidad, and M.~V. Escandell~Vidal.
\newblock 2023.
\newblock Using linguistic knowledge for automated text identification.
\newblock In {\em Proceedings of IberLEF 2023}. CEUR Workshop Proceedings.

\bibitem[\protect\citename{Azoulay}2024]{unesco-llm-bias}
Azoulay, A.
\newblock 2024.
\newblock Generative ai: Unesco study reveals alarming evidence of regressive gender stereotypes, March.

\bibitem[\protect\citename{Bublak}2024]{llm-risks-bublak}
Bublak, D.~D.
\newblock 2024.
\newblock Test it: Ai -vs- human generated material, April.

\bibitem[\protect\citename{Chowdhery \bgroup et al.\egroup }2022]{chowdhery2022palm}
Chowdhery, A., S.~Narang, J.~Devlin, M.~Bosma, G.~Mishra, A.~Roberts, P.~Barham, H.~W. Chung, C.~Sutton, S.~Gehrmann, P.~Schuh, K.~Shi, S.~Tsvyashchenko, J.~Maynez, A.~Rao, P.~Barnes, Y.~Tay, N.~Shazeer, V.~Prabhakaran, E.~Reif, N.~Du, B.~Hutchinson, R.~Pope, J.~Bradbury, J.~Austin, M.~Isard, G.~Gur-Ari, P.~Yin, T.~Duke, A.~Levskaya, S.~Ghemawat, S.~Dev, H.~Michalewski, X.~Garcia, V.~Misra, K.~Robinson, L.~Fedus, D.~Zhou, D.~Ippolito, D.~Luan, H.~Lim, B.~Zoph, A.~Spiridonov, R.~Sepassi, D.~Dohan, S.~Agrawal, M.~Omernick, A.~M. Dai, T.~S. Pillai, M.~Pellat, A.~Lewkowycz, E.~Moreira, R.~Child, O.~Polozov, K.~Lee, Z.~Zhou, X.~Wang, B.~Saeta, M.~Diaz, O.~Firat, M.~Catasta, J.~Wei, K.~Meier-Hellstern, D.~Eck, J.~Dean, S.~Petrov, and N.~Fiedel.
\newblock 2022.
\newblock Palm: Scaling language modeling with pathways.

\bibitem[\protect\citename{García and Segura-Bedmar}2024]{jor_isa_uc3m}
García, J.~F. and I.~Segura-Bedmar.
\newblock 2024.
\newblock Human after all: Using transformer based models to identify automatically generated text.
\newblock In {\em Proceedings of IberLEF 2024}. CEUR Workshop Proceedings.

\bibitem[\protect\citename{Gehrmann, Strobelt, and Rush}2019a]{gehrmann-etal-2019-gltr}
Gehrmann, S., H.~Strobelt, and A.~Rush.
\newblock 2019a.
\newblock {GLTR}: Statistical detection and visualization of generated text.
\newblock In M.~R. Costa-juss{\`a} and E.~Alfonseca, editors, {\em Proceedings of the 57th Annual Meeting of the Association for Computational Linguistics: System Demonstrations}, pages 111--116, Florence, Italy, July. Association for Computational Linguistics.

\bibitem[\protect\citename{Gehrmann, Strobelt, and Rush}2019b]{DBLP:journals/corr/abs-1906-04043}
Gehrmann, S., H.~Strobelt, and A.~M. Rush.
\newblock 2019b.
\newblock {GLTR:} statistical detection and visualization of generated text.
\newblock {\em CoRR}, abs/1906.04043.

\bibitem[\protect\citename{Heldt}2023]{llm-risks-heldt}
Heldt, J.
\newblock 2023.
\newblock The dangers of ai-generated content, October.

\bibitem[\protect\citename{Huang \bgroup et al.\egroup }2023]{huang2023survey}
Huang, L., W.~Yu, W.~Ma, W.~Zhong, Z.~Feng, H.~Wang, Q.~Chen, W.~Peng, X.~Feng, B.~Qin, and T.~Liu.
\newblock 2023.
\newblock A survey on hallucination in large language models: Principles, taxonomy, challenges, and open questions.

\bibitem[\protect\citename{Jim{\'e}nez-Zafra and Rangel}2023]{jimenez2023overview}
Jim{\'e}nez-Zafra, S.~M. and F.~Rangel.
\newblock 2023.
\newblock Overview of iberlef 2023: Natural language processing challenges for spanish and other iberian languages.

\bibitem[\protect\citename{Martínez \bgroup et al.\egroup }2023]{martínez2023combining}
Martínez, G., L.~Watson, P.~Reviriego, J.~A. Hernández, M.~Juarez, and R.~Sarkar.
\newblock 2023.
\newblock Combining generative artificial intelligence (ai) and the internet: Heading towards evolution or degradation?

\bibitem[\protect\citename{Obregon and Carrera}2023]{datificate}
Obregon, J. and B.~Carrera.
\newblock 2023.
\newblock Gpt2-small-spanish: a language model for spanish text generation (and more nlp tasks...).

\bibitem[\protect\citename{OpenAI \bgroup et al.\egroup }2024]{openai2024gpt4}
OpenAI, J.~Achiam, S.~Adler, S.~Agarwal, L.~Ahmad, I.~Akkaya, F.~L. Aleman, D.~Almeida, J.~Altenschmidt, S.~Altman, S.~Anadkat, R.~Avila, I.~Babuschkin, S.~Balaji, V.~Balcom, P.~Baltescu, H.~Bao, M.~Bavarian, J.~Belgum, I.~Bello, J.~Berdine, G.~Bernadett-Shapiro, C.~Berner, L.~Bogdonoff, O.~Boiko, M.~Boyd, A.-L. Brakman, G.~Brockman, T.~Brooks, M.~Brundage, K.~Button, T.~Cai, R.~Campbell, A.~Cann, B.~Carey, C.~Carlson, R.~Carmichael, B.~Chan, C.~Chang, F.~Chantzis, D.~Chen, S.~Chen, R.~Chen, J.~Chen, M.~Chen, B.~Chess, C.~Cho, C.~Chu, H.~W. Chung, D.~Cummings, J.~Currier, Y.~Dai, C.~Decareaux, T.~Degry, N.~Deutsch, D.~Deville, A.~Dhar, D.~Dohan, S.~Dowling, S.~Dunning, A.~Ecoffet, A.~Eleti, T.~Eloundou, D.~Farhi, L.~Fedus, N.~Felix, S.~P. Fishman, J.~Forte, I.~Fulford, L.~Gao, E.~Georges, C.~Gibson, V.~Goel, T.~Gogineni, G.~Goh, R.~Gontijo-Lopes, J.~Gordon, M.~Grafstein, S.~Gray, R.~Greene, J.~Gross, S.~S. Gu, Y.~Guo, C.~Hallacy, J.~Han, J.~Harris, Y.~He, M.~Heaton, J.~Heidecke, C.~Hesse, A.~Hickey,
  W.~Hickey, P.~Hoeschele, B.~Houghton, K.~Hsu, S.~Hu, X.~Hu, J.~Huizinga, S.~Jain, S.~Jain, J.~Jang, A.~Jiang, R.~Jiang, H.~Jin, D.~Jin, S.~Jomoto, B.~Jonn, H.~Jun, T.~Kaftan, Łukasz Kaiser, A.~Kamali, I.~Kanitscheider, N.~S. Keskar, T.~Khan, L.~Kilpatrick, J.~W. Kim, C.~Kim, Y.~Kim, J.~H. Kirchner, J.~Kiros, M.~Knight, D.~Kokotajlo, Łukasz Kondraciuk, A.~Kondrich, A.~Konstantinidis, K.~Kosic, G.~Krueger, V.~Kuo, M.~Lampe, I.~Lan, T.~Lee, J.~Leike, J.~Leung, D.~Levy, C.~M. Li, R.~Lim, M.~Lin, S.~Lin, M.~Litwin, T.~Lopez, R.~Lowe, P.~Lue, A.~Makanju, K.~Malfacini, S.~Manning, T.~Markov, Y.~Markovski, B.~Martin, K.~Mayer, A.~Mayne, B.~McGrew, S.~M. McKinney, C.~McLeavey, P.~McMillan, J.~McNeil, D.~Medina, A.~Mehta, J.~Menick, L.~Metz, A.~Mishchenko, P.~Mishkin, V.~Monaco, E.~Morikawa, D.~Mossing, T.~Mu, M.~Murati, O.~Murk, D.~Mély, A.~Nair, R.~Nakano, R.~Nayak, A.~Neelakantan, R.~Ngo, H.~Noh, L.~Ouyang, C.~O'Keefe, J.~Pachocki, A.~Paino, J.~Palermo, A.~Pantuliano, G.~Parascandolo, J.~Parish, E.~Parparita,
  A.~Passos, M.~Pavlov, A.~Peng, A.~Perelman, F.~de~Avila Belbute~Peres, M.~Petrov, H.~P. de~Oliveira~Pinto, Michael, Pokorny, M.~Pokrass, V.~H. Pong, T.~Powell, A.~Power, B.~Power, E.~Proehl, R.~Puri, A.~Radford, J.~Rae, A.~Ramesh, C.~Raymond, F.~Real, K.~Rimbach, C.~Ross, B.~Rotsted, H.~Roussez, N.~Ryder, M.~Saltarelli, T.~Sanders, S.~Santurkar, G.~Sastry, H.~Schmidt, D.~Schnurr, J.~Schulman, D.~Selsam, K.~Sheppard, T.~Sherbakov, J.~Shieh, S.~Shoker, P.~Shyam, S.~Sidor, E.~Sigler, M.~Simens, J.~Sitkin, K.~Slama, I.~Sohl, B.~Sokolowsky, Y.~Song, N.~Staudacher, F.~P. Such, N.~Summers, I.~Sutskever, J.~Tang, N.~Tezak, M.~B. Thompson, P.~Tillet, A.~Tootoonchian, E.~Tseng, P.~Tuggle, N.~Turley, J.~Tworek, J.~F.~C. Uribe, A.~Vallone, A.~Vijayvergiya, C.~Voss, C.~Wainwright, J.~J. Wang, A.~Wang, B.~Wang, J.~Ward, J.~Wei, C.~Weinmann, A.~Welihinda, P.~Welinder, J.~Weng, L.~Weng, M.~Wiethoff, D.~Willner, C.~Winter, S.~Wolrich, H.~Wong, L.~Workman, S.~Wu, J.~Wu, M.~Wu, K.~Xiao, T.~Xu, S.~Yoo, K.~Yu, Q.~Yuan,
  W.~Zaremba, R.~Zellers, C.~Zhang, M.~Zhang, S.~Zhao, T.~Zheng, J.~Zhuang, W.~Zhuk, and B.~Zoph.
\newblock 2024.
\newblock Gpt-4 technical report.

\bibitem[\protect\citename{Przybyla, Duran-Silva, and Egea-Gómez}2023]{taln-upf2023bilstm}
Przybyla, P., N.~Duran-Silva, and S.~Egea-Gómez.
\newblock 2023.
\newblock I’ve seen things you machines wouldn’t believe: Measuring content predictability to identify automatically-generated text.
\newblock In {\em Proceedings of IberLEF 2023}. CEUR Workshop Proceedings.

\bibitem[\protect\citename{Radford \bgroup et al.\egroup }2019]{radford2019language}
Radford, A., J.~Wu, R.~Child, D.~Luan, D.~Amodei, and I.~Sutskever.
\newblock 2019.
\newblock Language models are unsupervised multitask learners.

\bibitem[\protect\citename{Sarvazyan \bgroup et al.\egroup }2023]{autextification}
Sarvazyan, A.~M., J.~{\'A}. Gonz{\'a}lez, M.~Franco~Salvador, F.~Rangel, B.~Chulvi, and P.~Rosso.
\newblock 2023.
\newblock Overview of autextification at iberlef 2023: Detection and attribution of machine-generated text in multiple domains.
\newblock In {\em Procesamiento del Lenguaje Natural}, Jaén, Spain, September.

\bibitem[\protect\citename{Sarvazyan \bgroup et al.\egroup }2024]{PLN6628}
Sarvazyan, A.~M., J.~Ángel González, F.~Rangel, P.~Rosso, and M.~Franco-Salvador.
\newblock 2024.
\newblock Overview of iberautextification at iberlef 2024: Detection and attribution of machine-generated text on languages of the iberian peninsula.
\newblock {\em Procesamiento del Lenguaje Natural}, 73(0):421--434.

\bibitem[\protect\citename{Team}2023]{llm-risks-upwork}
Team, T.~U.
\newblock 2023.
\newblock The risks of ai-generated content: What you need to know, October.

\bibitem[\protect\citename{Tian and Cui}2023]{tian2023gptzero}
Tian, E. and A.~Cui.
\newblock 2023.
\newblock Gptzero: Towards detection of ai-generated text using zero-shot and supervised methods.

\bibitem[\protect\citename{Workshop \bgroup et al.\egroup }2023]{workshop2023bloom}
Workshop, B., :, T.~L. Scao, A.~Fan, C.~Akiki, E.~Pavlick, S.~Ilić, D.~Hesslow, R.~Castagné, A.~S. Luccioni, F.~Yvon, M.~Gallé, J.~Tow, A.~M. Rush, S.~Biderman, A.~Webson, P.~S. Ammanamanchi, T.~Wang, B.~Sagot, N.~Muennighoff, A.~V. del Moral, O.~Ruwase, R.~Bawden, S.~Bekman, A.~McMillan-Major, I.~Beltagy, H.~Nguyen, L.~Saulnier, S.~Tan, P.~O. Suarez, V.~Sanh, H.~Laurençon, Y.~Jernite, J.~Launay, M.~Mitchell, C.~Raffel, A.~Gokaslan, A.~Simhi, A.~Soroa, A.~F. Aji, A.~Alfassy, A.~Rogers, A.~K. Nitzav, C.~Xu, C.~Mou, C.~Emezue, C.~Klamm, C.~Leong, D.~van Strien, D.~I. Adelani, D.~Radev, E.~G. Ponferrada, E.~Levkovizh, E.~Kim, E.~B. Natan, F.~D. Toni, G.~Dupont, G.~Kruszewski, G.~Pistilli, H.~Elsahar, H.~Benyamina, H.~Tran, I.~Yu, I.~Abdulmumin, I.~Johnson, I.~Gonzalez-Dios, J.~de~la Rosa, J.~Chim, J.~Dodge, J.~Zhu, J.~Chang, J.~Frohberg, J.~Tobing, J.~Bhattacharjee, K.~Almubarak, K.~Chen, K.~Lo, L.~V. Werra, L.~Weber, L.~Phan, L.~B. allal, L.~Tanguy, M.~Dey, M.~R. Muñoz, M.~Masoud, M.~Grandury, M.~Šaško,
  M.~Huang, M.~Coavoux, M.~Singh, M.~T.-J. Jiang, M.~C. Vu, M.~A. Jauhar, M.~Ghaleb, N.~Subramani, N.~Kassner, N.~Khamis, O.~Nguyen, O.~Espejel, O.~de~Gibert, P.~Villegas, P.~Henderson, P.~Colombo, P.~Amuok, Q.~Lhoest, R.~Harliman, R.~Bommasani, R.~L. López, R.~Ribeiro, S.~Osei, S.~Pyysalo, S.~Nagel, S.~Bose, S.~H. Muhammad, S.~Sharma, S.~Longpre, S.~Nikpoor, S.~Silberberg, S.~Pai, S.~Zink, T.~T. Torrent, T.~Schick, T.~Thrush, V.~Danchev, V.~Nikoulina, V.~Laippala, V.~Lepercq, V.~Prabhu, Z.~Alyafeai, Z.~Talat, A.~Raja, B.~Heinzerling, C.~Si, D.~E. Taşar, E.~Salesky, S.~J. Mielke, W.~Y. Lee, A.~Sharma, A.~Santilli, A.~Chaffin, A.~Stiegler, D.~Datta, E.~Szczechla, G.~Chhablani, H.~Wang, H.~Pandey, H.~Strobelt, J.~A. Fries, J.~Rozen, L.~Gao, L.~Sutawika, M.~S. Bari, M.~S. Al-shaibani, M.~Manica, N.~Nayak, R.~Teehan, S.~Albanie, S.~Shen, S.~Ben-David, S.~H. Bach, T.~Kim, T.~Bers, T.~Fevry, T.~Neeraj, U.~Thakker, V.~Raunak, X.~Tang, Z.-X. Yong, Z.~Sun, S.~Brody, Y.~Uri, H.~Tojarieh, A.~Roberts, H.~W. Chung,
  J.~Tae, J.~Phang, O.~Press, C.~Li, D.~Narayanan, H.~Bourfoune, J.~Casper, J.~Rasley, M.~Ryabinin, M.~Mishra, M.~Zhang, M.~Shoeybi, M.~Peyrounette, N.~Patry, N.~Tazi, O.~Sanseviero, P.~von Platen, P.~Cornette, P.~F. Lavallée, R.~Lacroix, S.~Rajbhandari, S.~Gandhi, S.~Smith, S.~Requena, S.~Patil, T.~Dettmers, A.~Baruwa, A.~Singh, A.~Cheveleva, A.-L. Ligozat, A.~Subramonian, A.~Névéol, C.~Lovering, D.~Garrette, D.~Tunuguntla, E.~Reiter, E.~Taktasheva, E.~Voloshina, E.~Bogdanov, G.~I. Winata, H.~Schoelkopf, J.-C. Kalo, J.~Novikova, J.~Z. Forde, J.~Clive, J.~Kasai, K.~Kawamura, L.~Hazan, M.~Carpuat, M.~Clinciu, N.~Kim, N.~Cheng, O.~Serikov, O.~Antverg, O.~van~der Wal, R.~Zhang, R.~Zhang, S.~Gehrmann, S.~Mirkin, S.~Pais, T.~Shavrina, T.~Scialom, T.~Yun, T.~Limisiewicz, V.~Rieser, V.~Protasov, V.~Mikhailov, Y.~Pruksachatkun, Y.~Belinkov, Z.~Bamberger, Z.~Kasner, A.~Rueda, A.~Pestana, A.~Feizpour, A.~Khan, A.~Faranak, A.~Santos, A.~Hevia, A.~Unldreaj, A.~Aghagol, A.~Abdollahi, A.~Tammour, A.~HajiHosseini,
  B.~Behroozi, B.~Ajibade, B.~Saxena, C.~M. Ferrandis, D.~McDuff, D.~Contractor, D.~Lansky, D.~David, D.~Kiela, D.~A. Nguyen, E.~Tan, E.~Baylor, E.~Ozoani, F.~Mirza, F.~Ononiwu, H.~Rezanejad, H.~Jones, I.~Bhattacharya, I.~Solaiman, I.~Sedenko, I.~Nejadgholi, J.~Passmore, J.~Seltzer, J.~B. Sanz, L.~Dutra, M.~Samagaio, M.~Elbadri, M.~Mieskes, M.~Gerchick, M.~Akinlolu, M.~McKenna, M.~Qiu, M.~Ghauri, M.~Burynok, N.~Abrar, N.~Rajani, N.~Elkott, N.~Fahmy, O.~Samuel, R.~An, R.~Kromann, R.~Hao, S.~Alizadeh, S.~Shubber, S.~Wang, S.~Roy, S.~Viguier, T.~Le, T.~Oyebade, T.~Le, Y.~Yang, Z.~Nguyen, A.~R. Kashyap, A.~Palasciano, A.~Callahan, A.~Shukla, A.~Miranda-Escalada, A.~Singh, B.~Beilharz, B.~Wang, C.~Brito, C.~Zhou, C.~Jain, C.~Xu, C.~Fourrier, D.~L. Periñán, D.~Molano, D.~Yu, E.~Manjavacas, F.~Barth, F.~Fuhrimann, G.~Altay, G.~Bayrak, G.~Burns, H.~U. Vrabec, I.~Bello, I.~Dash, J.~Kang, J.~Giorgi, J.~Golde, J.~D. Posada, K.~R. Sivaraman, L.~Bulchandani, L.~Liu, L.~Shinzato, M.~H. de~Bykhovetz, M.~Takeuchi,
  M.~Pàmies, M.~A. Castillo, M.~Nezhurina, M.~Sänger, M.~Samwald, M.~Cullan, M.~Weinberg, M.~D. Wolf, M.~Mihaljcic, M.~Liu, M.~Freidank, M.~Kang, N.~Seelam, N.~Dahlberg, N.~M. Broad, N.~Muellner, P.~Fung, P.~Haller, R.~Chandrasekhar, R.~Eisenberg, R.~Martin, R.~Canalli, R.~Su, R.~Su, S.~Cahyawijaya, S.~Garda, S.~S. Deshmukh, S.~Mishra, S.~Kiblawi, S.~Ott, S.~Sang-aroonsiri, S.~Kumar, S.~Schweter, S.~Bharati, T.~Laud, T.~Gigant, T.~Kainuma, W.~Kusa, Y.~Labrak, Y.~S. Bajaj, Y.~Venkatraman, Y.~Xu, Y.~Xu, Y.~Xu, Z.~Tan, Z.~Xie, Z.~Ye, M.~Bras, Y.~Belkada, and T.~Wolf.
\newblock 2023.
\newblock Bloom: A 176b-parameter open-access multilingual language model.

\end{thebibliography}

\end{document}